\documentclass[11pt]{article}

\usepackage[preprint]{acl}

\usepackage{times}
\usepackage{latexsym}

\usepackage[T1]{fontenc}

\usepackage[utf8]{inputenc}

\usepackage{microtype}

\usepackage{inconsolata}

\usepackage{graphicx}

\usepackage{fontawesome5}
\usepackage[table]{xcolor}
\usepackage{booktabs}
\usepackage{tikz}
\usepackage{pgfplots}
\usetikzlibrary{patterns}
\usepackage{listings}
\lstdefinestyle{tofucode}{%
  basicstyle=\ttfamily\scriptsize,
  breaklines=true,
  breakatwhitespace=false,
  columns=fullflexible,
  keepspaces=true,
  showstringspaces=false,
  frame=single,
  framesep=5pt,
  xleftmargin=5pt,
  xrightmargin=5pt,
  rulecolor=\color{black!25},
  backgroundcolor=\color{black!2},
  aboveskip=4pt,
  belowskip=2pt,
}

\title{
\raisebox{-0.15cm}{\includegraphics[height=1.2em]{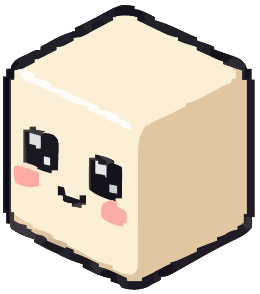}} 
\textsc{ToFu}: A White-Box, Token-Efficient Agent Harness for Researchers
}

\author{Junhao Ruan$^{1,2}$\thanks{Equal contribution.}, Yuan Ge$^{1*}$, Bei Li$^2$\thanks{Project leader.}, Yongjing Yin$^{2}$, Yuchun Fan$^{1,2}$, Xin Chen$^{2}$, \\\textbf{Jingang Wang$^{2}$,} 
\textbf{Chenglong Wang$^{1,3}$, Jingbo Zhu$^{1,3}$, Tong Xiao$^{1,3}$}\thanks{Corresponding author.} \\
$^1$ Northeastern University, China \\
$^2$ LongCat RSI, Meituan\\
$^3$ NiuTrans Research \\ 
}

\begin{document}
\maketitle
\begin{abstract}
Agentic coding tools present new opportunities to transform research workflows.
The performance of agent systems built depends on both large language models (LLMs) and the \textit{harness} around
LLMs, which is the orchestration code that determines an agent's behavior.
We present \textsc{ToFu}, an agentic harness for researchers that reads your codebase, edits files, runs commands, and integrates with your development tools. 
\textsc{ToFu} plays a dual role in research.
As a \textit{research assistant}, it supports practical research workflows with superior token efficiency, lower cost, and multilingual capability compared with existing agentic harnesses.
Its release under the \texttt{MIT License} further enables local deployment for privacy-sensitive users.
As a \textit{research object}, \textsc{ToFu} provides a white-box agentic harness that allows researchers to inspect, modify, and evaluate its orchestration logic, tool-use behavior, and harness design, while retaining strong benchmark performance and an application-level user experience.

{
    \small
    \makebox[1em][l]{\faGithub}\hspace{0.6em}  \href{https://github.com/NiuTrans/ToFu}{\texttt{github.com/NiuTrans/ToFu}} \\
}
\end{abstract}

\section{Introduction}

Large language models (LLMs) present new opportunities to transform intelligent assistants. Rather than merely receiving user requests and generating textual responses \citep{liu2024deepseek, singh2025openai, anthropic2024claude}, modern intelligent assistants can interact with external tools and environments to perform actions such as web search, file editing, and code execution \citep{yao2023react, anthropic2025claudecode, openai2025codex, li2025websailor}. 



\begin{figure}[tbp]
    \centering
    \centering
    \includegraphics[width=\columnwidth]{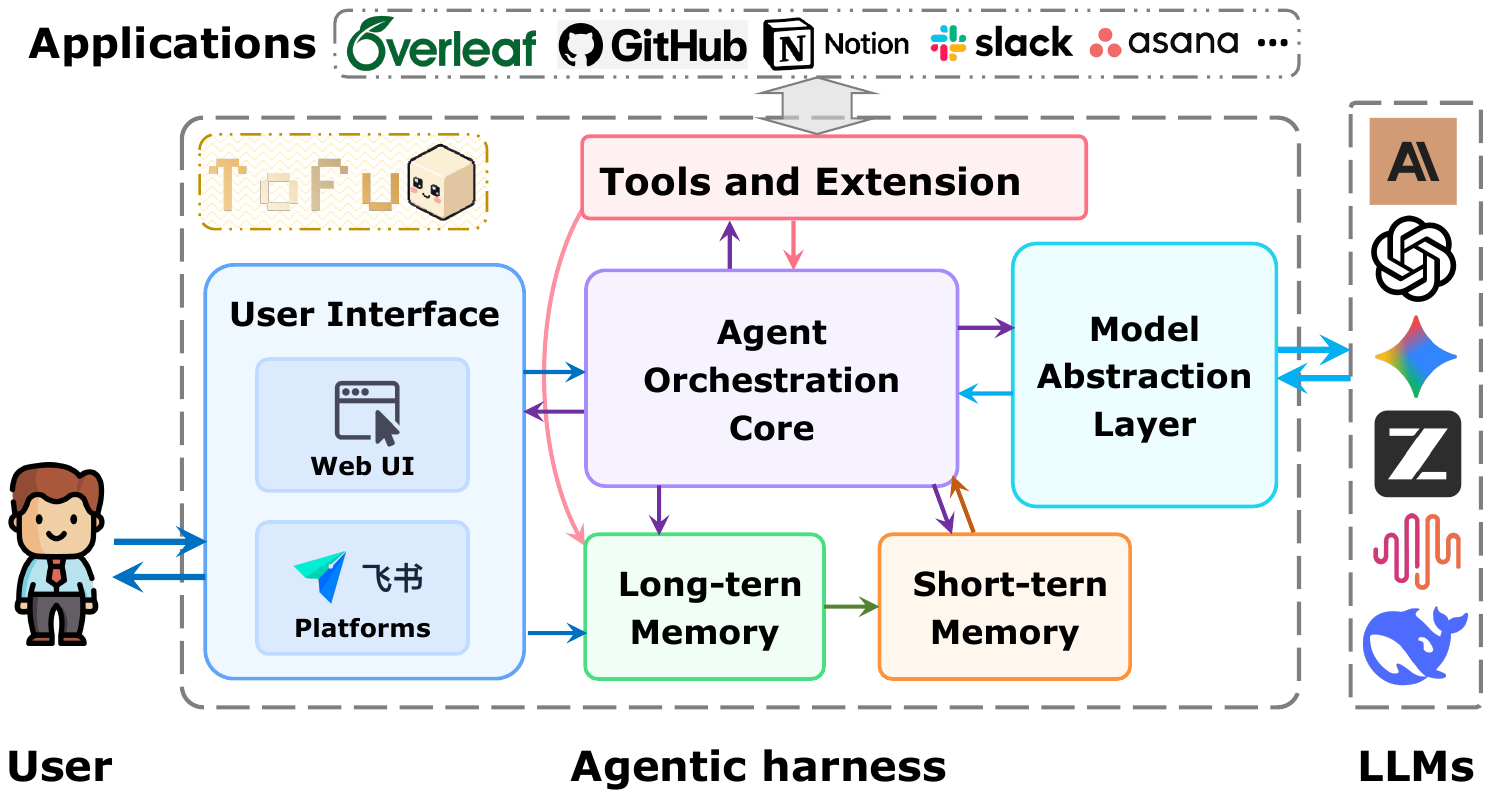}
    \caption{Human interaction with \textsc{ToFu} harness.}
    \label{fig:demo}
\end{figure}

The performance of agentic systems depends on both LLMs and the surrounding \textit{harness}: the orchestration code that shapes an agent's behavior.
An agentic system acts toward a goal with a degree of autonomy \citep{anthropic2025claudecode, openai2025codex, openclaw2026}. Harness enables the agent to read documents or codebase, plan sequences of actions, execute them using real development tools, evaluate the results, and adjust its approach.

The widespread adoption of current agentic systems is constrained by the following challenges of agentic harness. (i) \textit{Token efficiency}: even responding to a simple query such as “hello” requires an agentic system to consume thousands of context tokens, while complex multi-step tasks incur substantial financial costs. (ii) \textit{Multilingual capability}: the multilingual performance of agentic systems depends on that of backbone LLMs, resulting in a better user experience for English speakers than for users of other widely spoken or low-resource languages \citep{shi2023language, qin2024multilingual, mu2024revealing, hofman2026maps}. (iii) \textit{Open-source projects and local deployment}: privacy-sensitive users require both LLMs and agentic harness to be deployed locally to ensure information security \citep{xu2024device, ge2024clustering, siyan2025papillon}.
(iv) \textit{White-box research instrument}: current harnesses face the trade-off: research-oriented harnesses are reproducible but benchmark-shaped and lack a usable product surface \citep{sweagent2024, wang2025openhands}, while polished products default to proprietary models that are silently updated server-side, making them poor instruments for research \citep{openai2025codex, anthropic2025claudecode}. Research on harness evolution \citep{lee2026meta, seong2026last, lin2026agentic, lin2026harness, chen2026harnessforge} requires an open-source harness with both strong benchmark performance and application-level user experience.

In this paper, we present \textsc{ToFu}, a token-efficient, multilingual-enhanced, and open-source agentic harness for researchers.
Through three-layer context compaction, \textsc{ToFu} achieves leading performance with superior token efficiency. Its multilingual and swarm-enhanced modes further expand the capability boundaries of agentic systems. Moreover, \textsc{ToFu} supports a wide range of applications via Model Context Protocols (MCPs), enabling direct interaction with platforms such as Overleaf, GitHub, and Notion, as well as the use of agentic coding tools such as Claude Code as callable tools. More importantly, \textsc{ToFu} serves as an open-source baseline that combines a usable user experience with leading performance.
Our contributions are:

\begin{figure*}[tbp]
    \centering
    \centering
    \includegraphics[width=2\columnwidth]{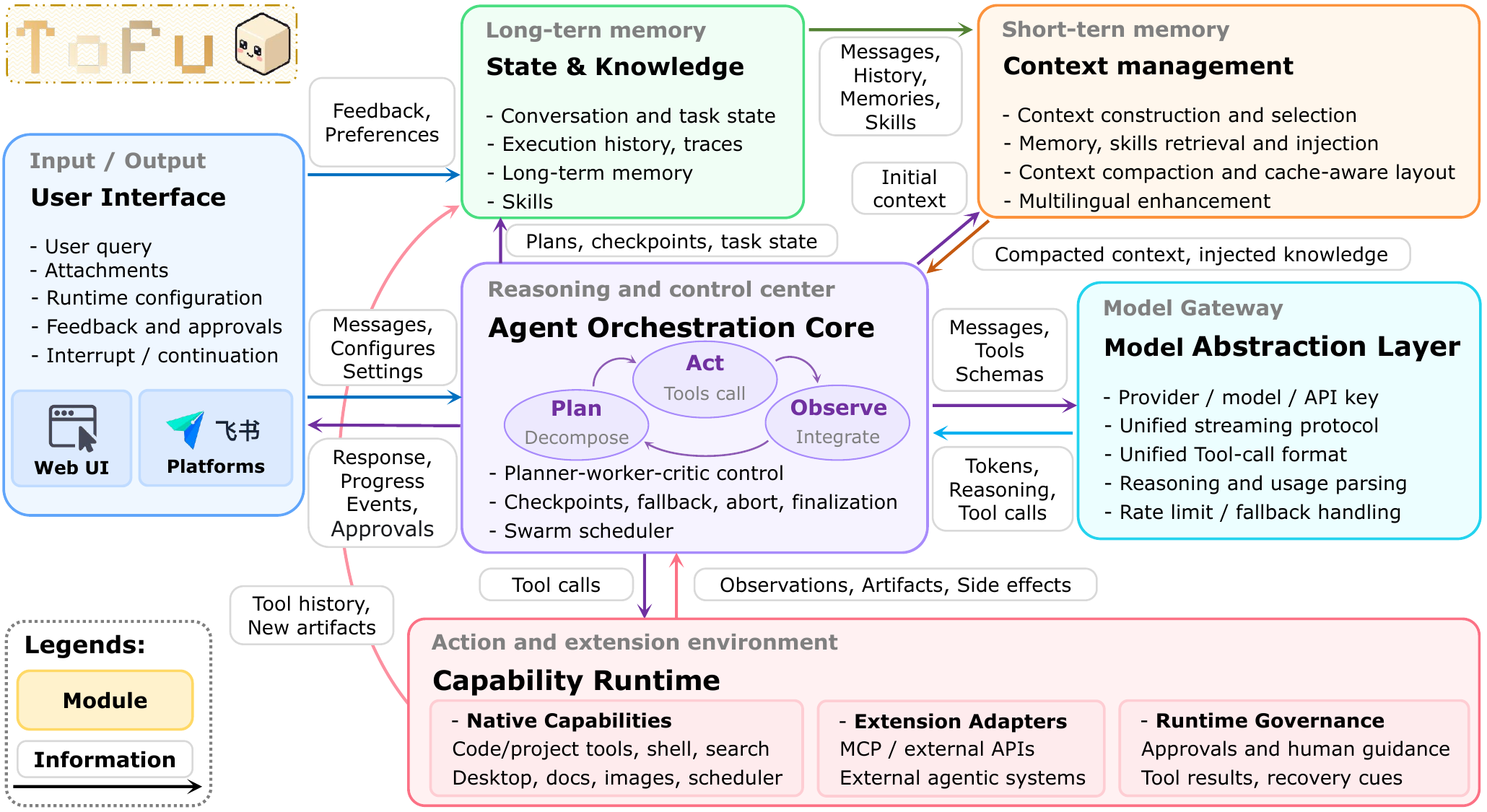}
    \caption{The overview of \textsc{ToFu} harness.}
    \label{fig:overview}
\end{figure*}

\begin{itemize}
    \item We present \textsc{ToFu}, an open-source agentic harness that extends the boundary of AI research assistants from reactively providing advice to automatically finishing goals.

    \item We employ specific harness design and context management to improve token efficiency. On \texttt{SWE-bench Verified}, \textsc{ToFu} uses \textit{28.4\% fewer tokens on average} (up to \textit{43.6\%}) than \textsc{Claude Code} across three backbone LLMs, while delivering better performance.


    \item We enhance \textsc{ToFu} via a multilingual mode, where an auxiliary translation layer enables the agent to operate internally in English. This exploits the performance advantages of English centric LLMs, while the source language remains for external interface with the user.

    \item The open-source nature of \textsc{ToFu} enables users to further extend and customize it for task-specific needs, while its white-box design offers a baseline for harness research that balances effectiveness with user experience.
    
\end{itemize}


\section{\textsc{ToFu} architecture}

\subsection{Framework overview}
This section introduces the abstract modular architecture of \textsc{ToFu}, and the information exchange among them. 
\textsc{ToFu} consists of six modules: the User Interface, Agent Orchestration Core, Model Abstraction Layer, Capability Runtime, State \& Knowledge, and Context Management.

\paragraph{User Interface} defines the interaction between users and \textsc{ToFu}. User can use \textsc{ToFu} through a Web UI or a social platform bot. Similar to interactions with LLMs, users can submit queries and attachments, including files, screenshots, and file paths. In addition, the interface allows users to configure settings such as model selection, thinking mode, and whether to enable sub-agents or multilingual enhancement.
Users can interrupt the agentic tool during its reasoning process or resume the conversation from a checkpoint. For safety considerations, \textsc{ToFu} requires explicit user approval before executing actions that may be risky, irreversible, or involve external permissions. The interface sends the final user message and configuration settings to the Agent Orchestration Core for processing, and also stores preference information, such as feedback and approvals, in long-term memory.

\paragraph{Agent Orchestration Core} serves as the agent control center of \textsc{ToFu}. It is not the model itself, but rather a control layer responsible for organizing agent behaviors. 
Orchestration core decomposes user requests, coordinates planner-worker-critic style reasoning, manages tool calls, tracks intermediate observations, and controls fallback, checkpointing, interruption, and finalization. 
In general, the task of the Orchestration Core is to receive, integrate, and dispatch information. It organizes short- and long-term memory through the State \& Knowledge and Context Management modules, enables tool invocation and external system interaction through the Capability Runtime module, and accesses LLMs through the Model Abstraction Layer module.
For complex tasks, it also invokes the swarm scheduler, enabling multiple sub-agents to work in parallel while preserving dependency-aware coordination and final synthesis.

\paragraph{Model Abstraction Layer} provides a unified gateway over heterogeneous LLM providers, models, API keys, and streaming protocols. It normalizes provider-specific differences into a consistent message and tool-call interface, parses reasoning traces and usage statistics, and handles rate limits, retries, and fallback routing. This abstraction allows the orchestration core to remain model-agnostic while still 
exploiting different models for reasoning, translation or summarization.

\paragraph{Capability Runtime} is the action and extension environment through which ToFu interacts with external systems. It is responsible for actually executing the tool call process and returning the results of the tool call along with any resulting effects. Capability Runtime module includes native capabilities such as code/project tools, shell execution, web search, desktop/browser access, document processing, image generation, and scheduling, as well as extension adapters such as MCPs and external agent APIs. \textsc{ToFu} supports applications such as Overleaf, GitHub, Slack and Notion through MCPs. Runtime governance mechanisms, including approval checks, human guidance, tool-result handling, and recovery cues, ensure that these capabilities remain controllable during execution.

\paragraph{State \& Knowledge} maintains the persistent information needed for continuity across tasks and sessions. It stores conversation state, execution history, task traces, long-term memories, and reusable skills, allowing ToFu to recover prior context without replaying entire conversations. This layer supports both operational state tracking and accumulated experiential knowledge, so future reasoning can reuse past decisions, project conventions, bug patterns, and workflow knowledge.

\paragraph{Context management} constructs the short-term working context that is actually sent to the model at each reasoning step. It selects and injects relevant messages, memories, skills, tool results, and knowledge; applies context compaction and cache-aware layout strategies; and supports multilingual enhancement detailed in section \ref{sec:m-enh}. The purpose of this module is to maximize useful information density while controlling token cost, prompt-cache stability, and context-window pressure.

\subsection{Three-layer context compaction}
\label{sec:compaction}
\textsc{ToFu} adopts a layered context compaction strategy to manage long agentic conversations under limited context windows. The design aims to reduce redundant or low-value tokens while preserving the information necessary for continued reasoning, tool use, and task completion. Instead of applying a single global truncation rule, ToFu combines three complementary forms of compaction.

First, \textsc{ToFu} performs size-aware budgeting for tool outputs. Large results from search, command execution, or web fetching are externalized and replaced with compact previews and recoverable references, so the model keeps awareness of the evidence without carrying the full text in context. File-reading outputs are treated more conservatively, since truncating source-code content often causes repeated tool calls and increases total cost.

Second, \textsc{ToFu} applies cache-aware micro-compaction to cold history. Recent tool results and reasoning traces are preserved verbatim, while older bulky tool outputs, stale reasoning blocks, and image-heavy results are replaced with concise placeholders. This process is deterministic and does not require an additional LLM call. It is also designed to avoid modifying cached prefixes, thereby reducing context size without unnecessarily invalidating prompt-cache reuse.

Third, when the conversation approaches the usable context limit, \textsc{ToFu} triggers query-aware semantic compaction. A lightweight model (e.g. \texttt{GPT-4o}) summarizes older turns with respect to the current user query: critical turns are preserved in high fidelity, useful turns are compressed, tangential turns are briefly mentioned, and irrelevant turns are removed. The current in-flight turn is always kept intact. This produces a compact working-state summary that retains user instructions, technical decisions, file context, errors, and pending steps, enabling the task to continue after compression with minimal loss of operational context.


\subsection{Memory and retrial sub-system}

The memory of \textsc{ToFu} implements as a persistent, retrieval-oriented knowledge layer rather than as raw conversation replay. Memories are stored as structured Markdown records with metadata such as name, description, tags, scope, and runtime eligibility. The system distinguishes project-specific memories from global memories, allowing reusable experience to be shared across tasks while still preserving project-local conventions. Instead of injecting all memory contents into every prompt, \textsc{ToFu} only provides a compact indication that memories are available, then the model retrieves relevant memories on demand.

Memory retrieval is based on lightweight BM25 ranking over memory metadata and content. This design avoids the cost and latency of dense embedding calls while remaining effective for technical memories, where exact terms such as library names, error messages, APIs, and file patterns are often highly informative. The model may search memories with task-specific keywords, inspect the returned memory paths, and then load only the records that are actually useful. Memories can also be created, updated, deleted, or merged during task execution, enabling the system to accumulate bug patterns, project conventions, user preferences, and workflow knowledge over time.

Overall, the memory and retrieval system gives \textsc{ToFu} both long-term experiential recall and up-to-date external evidence, while keeping context usage, latency, and retrieval noise under control.

\subsection{Multilingual enhancement}
\label{sec:m-enh}
ToFu supports multilingual interaction through a translate-then-reason workflow. When the user communicates in a non-English language, the input is first translated into English before being sent to the main reasoning model, allowing the system to benefit from stronger English-centric model capabilities and broader English web resources. After the model completes its reasoning and generates a response, the output is translated back into the user’s original language for display. The original text is preserved alongside the translated version, and code blocks or explicitly marked non-translatable spans are protected to avoid semantic or formatting corruption.

\subsection{Swarm scheduler}

\textsc{ToFu} uses a dependency-aware swarm scheduler to coordinate multiple sub-agents for complex tasks. Each sub-task is represented as an agent objective with optional dependencies, forming a directed acyclic graph. Independent agents are executed in parallel, while dependent agents are delayed until their prerequisite results become available. This allows \textsc{ToFu} to exploit parallelism without sacrificing task coherence.


The scheduler injects completed prerequisite results into downstream agents’ context, enabling later agents to build directly on earlier findings rather than rediscovering the same information. To improve robustness, agent execution is bounded by concurrency control, rate-limit-aware execution, and automatic retry with failure context. In reactive mode, a master agent periodically reviews completed results and decides whether the current evidence is sufficient or whether additional agents should be spawned. Thus, the swarm scheduler provides an adaptive execution mechanism that balances parallel speed, dependency correctness, and iterative task refinement.



\begin{table*}[t]
    \centering
    \small
    \begin{tabular}{lcccc}
    \toprule
    \textbf{Agentic Harness} & \textbf{Pass@1 (Solved/Attempted)} & \textbf{Avg. Cost (\$/inst.)} & \textbf{Avg. Tokens (/inst.)} & \textbf{Avg. Turns} \\ \midrule
    
    \multicolumn{5}{l}{\cellcolor{black!10} \textbf{\textit{I. Claude opus 4.6}} \citep{anthropic2026claude}} \\
    \textsc{ToFu} & \textbf{83.2\% (416/500)} & \$5.13 & \textbf{663,881} & \textbf{17.5} \\
    \textsc{Claude Code} & 79.6\% (398/500) & \textbf{\$4.97} & 837,764 & 18.6 \\
    \textsc{OpenCode} & 74.2\% (371/500) & \$7.00 & 757,925 & 22.7 \\
    \multicolumn{5}{l}{\cellcolor{black!10} \textbf{\textit{II. GLM 5.1}} \citep{zhipu2026glm}} \\
    \textsc{ToFu} & \textbf{80.4\% (402/500)} & \$1.19 & 575,255 & 17.5 \\
    \textsc{Claude Code} & 77.6\% (388/500) & \$1.67 & 1,020,453 & 20.0 \\
    \textsc{OpenCode} & 71.6\% (358/500) & \textbf{\$0.81} & \textbf{568,097} & \textbf{15.5} \\
    \multicolumn{5}{l}{\cellcolor{black!10} \textbf{\textit{III. DeepSeek-v4-pro}} \citep{deepseekai2026deepseekv4}} \\
    \textsc{ToFu} & \textbf{80.2\% (401/500)} & \$1.66 & 1,110,192 & 21.9 \\
    \textsc{Claude Code} & 75.2\% (376/500) & \$1.86 & 1,401,574 & 21.6 \\
    \textsc{OpenCode} & 71.8\% (359/500) & \textbf{\$1.08} & \textbf{831,294} & \textbf{17.6} \\ \bottomrule
    \end{tabular}
    \caption{Performance of \textsc{ToFu}, \textsc{Claude Code}, and \textsc{OpenCode} on the full \texttt{SWE-bench Verified}  across three agentic LLMs. All patches are graded by the official SWE-bench harness inside udocker.}
    \label{tab:swe}
\end{table*}

\section{\textsc{ToFu} application}

\subsection{User interface}
\textsc{ToFu}’s UI exposes agent behavior through a compact set of runtime controls rather than a plain chat interface. As shown in Fig. \ref{fig:interface} left, the bottom toolbar organizes these controls into four groups. Enhance enables auxiliary capabilities such as code execution, memory injection, and automatic translation. Tools grants access to external action environments, including browser control, desktop operation, scheduling, image generation, and human guidance. Mode selects higher-level orchestration strategies such as swarm-based parallel agents or autonomous plan-execute-review loops. Search controls whether the agent can retrieve external web information during a turn. The left sidebar provides persistent conversation management, allowing users to create new chats, browse and resume dialogue history, organize dialogues into folders, and perform actions such as referencing, moving, duplicating, or deleting a conversation. Dashboards at the top of the left sidebar are the paper reader, history search, daily report, and settings.


\subsection{Paper Reader}
As shown in Fig. \ref{fig:interface} right, \textsc{ToFu} integrates a paper reader feature. Uploading a PDF or pasting an arXiv URL opens a split-pane view (PDF left, chat right) with grounded Q\&A over the parsed text, a generated paper report, and a translated PDF paper via Babel PDF\footnote{\url{https://app.immersivetranslate.com/babel-doc}}.
The Q\&A feature and paper report generation is powered by \textsc{ToFu} harness.

\subsection{Overleaf}
\textsc{ToFu} can assist scientific manuscript writing in Overleaf by connecting the agent to the project workspace through our open-source \texttt{Overleaf MCP}\footnote{\url{https://github.com/rangehow/overleaf-mcp}}. This integration turns an Overleaf project into an editable and inspectable writing environment, where \textsc{ToFu} can list projects and files, read LaTeX sources, analyze document sections, revise abstracts or method descriptions, update specific sections, manage auxiliary files, inspect version history and diffs, and compile the manuscript to obtain PDFs and logs. In this way, \textsc{ToFu} supports the full academic writing loop, from understanding the current manuscript and proposing revisions to applying precise LaTeX edits and validating the compiled result within the user’s existing Overleaf.


\section{Evaluation}



Agentic coding is the primary task for harnesses since system-level interaction relies on understanding both problems and programming tools. 

\subsection{Evaluation setup}

\paragraph{Dataset}
We utilize \texttt{SWE-bench Verified} \citep{openai2024swebenchverified}, a human-filtered subset of 500 instances from SWE-bench \citep{jimenez2024swebench} created in collaboration with OpenAI. 
\texttt{SWE-bench} is designed to evaluate the ability of LLMs/agents to solve real-world software engineering tasks. It requires models to fix existing issues in large and complex GitHub repositories, similar to how software engineers operate.
Human annotators reviewed each instance to ensure the problem descriptions are clear, the test patches are correct, and the tasks are solvable given the available information.

\paragraph{Metrics}
We evaluate the agentic coding ability using \texttt{Pass@1}, indicating the reliability to solve the issue of a single try. Besides, we focus on the \textit{system efficiency} of agentic harnesses. We utilize \texttt{Avg. Cost(\$/inst.)} and \texttt{Avg. Tokens(\$/inst.)} to record the financial cost and total token usage for each instance in average, respectively. We further record the average turn, which includes both environment feedback and tool calls by \texttt{Avg. Turns}.

\paragraph{Agentic harnesses and Models} We compare \textsc{ToFu} with two baselines: commercial \textsc{Claude Code} \citep{anthropic2025claudecode} and open-source CLI OpenCode \citep{opencode2026}. 
As the performance of agentic systems depends on both \textit{design of agentic harhesses} and \textit{intelligence of backbone LLMs}, we evaluated three leading LLMs: \texttt{Claude opus 4.6} \citep{anthropic2026claude}, \texttt{GLM 5.1} \citep{zhipu2026glm}, and \texttt{DeepSeek-v4-pro} \citep{deepseekai2026deepseekv4}.

\subsection{Evaluation results}

\paragraph{\textsc{ToFu} outperforms \textsc{Claude Code} with fewer tokens.}
As shown in Table \ref{tab:swe}, \textsc{ToFu} achieves the best \texttt{Pass@1} on the \texttt{SWE-bench Verified} benchmark across all three backbone LLMs. 
Averaged over the three LLMs, \textsc{ToFu} outperforms \textsc{Claude Code} and \textsc{OpenCode} by 3.8 and 8.7 percentage points in \texttt{Pass@1}, respectively. 
More importantly, this improvement is achieved with substantially lower token consumption than \textsc{Claude Code}: \textsc{ToFu} uses fewer tokens across all LLMs, reducing the average token usage by 28.4\%. This suggests that the performance gain does not come from longer interactions or more extensive context usage, but from a more effective harness design. Although \textsc{OpenCode} is sometimes cheaper or more token-efficient, its solving performance is consistently lower than \textsc{ToFu}, indicating a clear trade-off between cost and task success. 
Note that \texttt{Avg. Tokens} calculates both input and output tokens, so fewer token usages may cost more because the output tokens are more expensive.
Overall, these results show that \textsc{ToFu} provides a stronger accuracy--efficiency balance, improving pass rate while maintaining competitive cost, token usage, and turns.


\paragraph{More computations do not mean better performance}
An interesting observation from the comparison between \textsc{ToFu} and \textsc{Claude Code} is that using more tokens for test-time scaling does not necessarily lead to better performance. On the contrary, \textsc{ToFu} outperforms \textsc{Claude Code} with fewer token usages and comparable costs. This suggests that \textit{there remains substantial room for improving the token efficiency of current harnesses}, which is one of our future optimization goals.

\subsection{Multilingual enhancement}


We utilize \texttt{MAPS} \citep{hofman2026maps}, a multilingual benchmark for agent performance and security. 
\texttt{MAPS} consists of four benchmarks: \texttt{GAIA}, \texttt{MATH}, \texttt{SWE-bench}, and \texttt{Agent Security}, and has been translated into ten languages: Spanish (es), German (de), Arabic (ar), Russian (ru), Japanese (ja), Portuguese (pt), Hindi (hi), Hebrew (he), Korean (ko), and Italian (it). We evaluate \textsc{ToFu}'s agentic coding performance via \texttt{MAPS:SWE-bench} on all languages utilizing \texttt{Claude opus 4.6}.

As shown in Fig. \ref{fig:multilingual}, the experimental results indicate that multilingual enhancement outperforms directly using the source language by an average of 2.5 accuracy points. Specifically, it yields improvements in 7 languages, while the remaining three are nearly unchanged or exhibit slight declines.



\definecolor{uyellow}{RGB}{253,186,107}
\definecolor{ured}{RGB}{235,096,070}
\definecolor{upurple}{RGB}{175,135,220} 
\definecolor{ublue}{RGB}{076,135,220}  

\pgfmathsetlengthmacro{\BarW}{8pt}
\pgfmathsetlengthmacro{\BarStep}{4.5pt}

\pgfmathsetlengthmacro{\Shiftwith}{  -\BarStep}
\pgfmathsetlengthmacro{\Shiftwo}{  \BarStep}

\tikzset{
  solidbar/.style={},
  hatchbar/.style={
    postaction={draw=none, pattern=north east lines}
  }
}

\begin{figure}[t]
    \centering
    \begin{tikzpicture}
    \tikzstyle{textonly} = [font=\small,align=left]
    \tikzstyle{sublayers} = [rectangle,draw,minimum width=0.6cm,rounded corners=0pt,align=center,inner sep=0pt,minimum height=0.28cm,font=\scriptsize]
    \begin{axis}[
        ybar,
        bar width=\BarW,
        width=1.12\linewidth,
        height=0.56\linewidth,
        ymin=30, ymax=78,
        ytick={0,20,40,50,60,70},
        symbolic x coords={Spanish,German,Arabic,Russian,Japanese,
        Portuguese,Hindi,Hebrew,Korean,Italian},
        xtick={Spanish,German,Arabic,Russian,Japanese,
        Portuguese,Hindi,Hebrew,Korean,Italian},
        xticklabels={Es,De,Ar,Ru,Ja,Pt,Hi,He,Ko,It},
        enlarge x limits=0.07,
        axis line style={thick},
        tick style={thick},
        label style={font=\normalsize},
        tick label style={font=\normalsize},
        ymajorgrids,
        grid style={dashed, gray!30},
        xtick pos=left,
        clip=false
    ]

    \addplot+[solidbar, hatchbar, draw=ured!80, fill=ured!40, pattern color=ured!80!black, bar shift=\Shiftwith] coordinates
        {(Spanish,63)};
    \addplot+[solidbar, hatchbar, draw=ublue!80, fill=ublue!40, pattern color=ublue!80!black, bar shift=\Shiftwo] coordinates
        {(Spanish,60)};

    \addplot+[solidbar, hatchbar, draw=ured!80, fill=ured!40, pattern color=ured!80!black, bar shift=\Shiftwith] coordinates
        {(German,65)};
    \addplot+[solidbar, hatchbar, draw=ublue!80, fill=ublue!40, pattern color=ublue!80!black, bar shift=\Shiftwo] coordinates
        {(German,59)};

    \addplot+[solidbar, hatchbar, draw=ured!80, fill=ured!40, pattern color=ured!80!black, bar shift=\Shiftwith] coordinates
        {(Arabic,60)};
    \addplot+[solidbar, hatchbar, draw=ublue!80, fill=ublue!40, pattern color=ublue!80!black, bar shift=\Shiftwo] coordinates
        {(Arabic,62)};

    \addplot+[solidbar, hatchbar, draw=ured!80, fill=ured!40, pattern color=ured!80!black, bar shift=\Shiftwith] coordinates
        {(Russian,62)};
    \addplot+[solidbar, hatchbar, draw=ublue!80, fill=ublue!40, pattern color=ublue!80!black, bar shift=\Shiftwo] coordinates
        {(Russian,63)};

    \addplot+[solidbar, hatchbar, draw=ured!80, fill=ured!40, pattern color=ured!80!black, bar shift=\Shiftwith] coordinates
        {(Japanese,63)};
    \addplot+[solidbar, hatchbar, draw=ublue!80, fill=ublue!40, pattern color=ublue!80!black, bar shift=\Shiftwo] coordinates
        {(Japanese,58)};

    \addplot+[solidbar, hatchbar, draw=ured!80, fill=ured!40, pattern color=ured!80!black, bar shift=\Shiftwith] coordinates
        {(Portuguese,63)};
    \addplot+[solidbar, hatchbar, draw=ublue!80, fill=ublue!40, pattern color=ublue!80!black, bar shift=\Shiftwo] coordinates
        {(Portuguese,57)};

    \addplot+[solidbar, hatchbar, draw=ured!80, fill=ured!40, pattern color=ured!80!black, bar shift=\Shiftwith] coordinates
        {(Hindi,66)};
    \addplot+[solidbar, hatchbar, draw=ublue!80, fill=ublue!40, pattern color=ublue!80!black, bar shift=\Shiftwo] coordinates
        {(Hindi,61)};

    \addplot+[solidbar, hatchbar, draw=ured!80, fill=ured!40, pattern color=ured!80!black, bar shift=\Shiftwith] coordinates
        {(Hebrew,60)};
    \addplot+[solidbar, hatchbar, draw=ublue!80, fill=ublue!40, pattern color=ublue!80!black, bar shift=\Shiftwo] coordinates
        {(Hebrew,64)};

    \addplot+[solidbar, hatchbar, draw=ured!80, fill=ured!40, pattern color=ured!80!black, bar shift=\Shiftwith] coordinates
        {(Korean,66)};
    \addplot+[solidbar, hatchbar, draw=ublue!80, fill=ublue!40, pattern color=ublue!80!black, bar shift=\Shiftwo] coordinates
        {(Korean,61)};

    \addplot+[solidbar, hatchbar, draw=ured!80, fill=ured!40, pattern color=ured!80!black, bar shift=\Shiftwith] coordinates
        {(Italian,64)};
    \addplot+[solidbar, hatchbar, draw=ublue!80, fill=ublue!40, pattern color=ublue!80!black, bar shift=\Shiftwo] coordinates
        {(Italian,61)};

    \end{axis}

    \node[sublayers, draw=ured!80, fill=ured!40, postaction={pattern=north east lines, pattern color=ured!80!black}]
      (tip) at (1.0,2.4) {};
    \node[textonly] at ([xshift=1.2cm]tip) {w/ M-Enh};
    
    \node[sublayers, draw=ublue!80, fill=ublue!40, postaction={pattern=north east lines, pattern color=ublue!80!black}]
          at ([xshift=3.0cm]tip) {};
    \node[textonly] at ([xshift=4.3cm]tip) {w/o M-Enh};

    \end{tikzpicture}
    
    \caption{Performance of \textsc{ToFu} with multilingual enhancement (M-Enh) on \texttt{MAPS: SWE-bench}.}
    \label{fig:multilingual}

\end{figure}
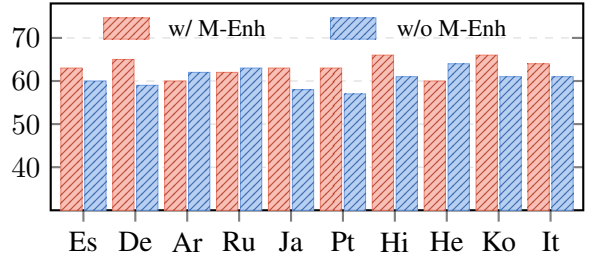




\section{Related work}



\paragraph{Agentic coding tools.}
An agentic system acts toward a goal with a degree of autonomy, rather than LLMs that respond to one prompt at a time or an agent workflow designed for specific tasks \citep{anthropic2025claudecode, openai2025codex, openclaw2026, maxclaw}. Agentic coding tools read a codebase, plan a sequence of actions, execute them using real development tools, evaluate the results, and adjust their approach.
The developer sets the objective and retains control over what gets committed, but the execution loop runs independently.

\paragraph{Agentic harnesses.}
Harness is as important as LLMs to decide the performance of agentic system. 
Therefore, harness evolution \citep{lee2026meta, seong2026last, lin2026agentic, lin2026harness, chen2026harnessforge} seeks better harnesses utilizing self-evolution or active learning. However, harness evolution requires an open-source harness with both strong benchmark performance and application-level user experience.
Current research-oriented harnesses are reproducible but benchmark-shaped, which means they are specific optimized on agentic coding benchmarks, and lack a usable product surface \citep{sweagent2024, wang2025openhands}. Meanwhile, commercial products such as \texttt{Codex} default to proprietary models making them poor

\section{Conclusion}
We present \textsc{ToFu}, an agentic harness for researchers with superior token efficiency, lower cost, and stronger multilingual capability than existing agentic harnesses.
We release \textsc{ToFu} under the \texttt{MIT License} to enable local deployment for privacy-sensitive users. 
Moreover, \textsc{ToFu} provides a white-box harness that retains strong benchmark performance and an application-level user experience.

\section*{Limitations}

This work has several limitations. First, our evaluation mainly focuses on coding ability, while evaluation in broader research-assistant scenarios is limited to a small human preference study with only three participants; due to cost constraints, we have not yet conducted large-scale studies with real users. Second, although \textsc{ToFu} achieves strong performance with lower token usage, we do not yet provide a systematic quantitative analysis of the trade-off between token efficiency and task performance, which we leave to future work.

\section*{Acknowledgments}
This work was supported in part by the National Science Foundation of China (Nos. 62276056 and U24A20334), the Yunnan Fundamental Research Projects (No.202401BC070021), the Yunnan Science and Technology Major Project (No. 202502AD080014), the Fundamental Research Funds for the Central Universities (Nos. N25BSS054 and N25BSS094), and the Program of Introducing Talents of Discipline to Universities, Plan 111 (No.B16009).

\bibliography{custom}
\clearpage

\appendix
\label{sec:appendix}

\section{Human Evaluation}
\label{sec:human-eval}

We evaluate on 100 tasks, spanning time-sensitive facts,
false-premise correction, citation honesty, multi-source comparison,
fact-checking, and documentation lookup. Both systems use the same model \texttt{Claude opus 4.6}.
Each task was rated blind by a panel of human annotators, in
position counterbalanced order, who scored accuracy, groundedness (claims
backed by independently checkable sources), and overall usability; accuracy
was scored against author-verified ground truth.

Across 700 pairwise comparisons, human raters prefer \textsc{ToFu} in
77\% of cases, versus 18\% for the baseline, with the remaining 5\%
rated as ties. Inter-annotator agreement is high, and every
annotator prefers \textsc{ToFu} overall (per-annotator win rate $76$--$86\%$).
\textsc{ToFu} leads on all three criteria, and the margins are statistically significant ($p<0.001$). The
preference holds in every category, including the hardest, documentation
lookup (58\%), indicating a genuine capability difference rather than an
artefact of task selection.

\begin{table}[h]
\centering\small
\begin{tabular}{lcc}
\toprule
Category & \textsc{ToFu} win & $n$ \\
\midrule
Multi-source comparison   & 97\% & 70 \\
False-premise correction  & 94\% & 84 \\
Time-sensitive facts      & 80\% & 112 \\
Common-knowledge          & 80\% & 112 \\
Citation honesty          & 79\% & 84 \\
Fact-checking             & 73\% & 70 \\
Documentation lookup      & 58\% & 168 \\
\midrule
\textbf{Overall}          & \textbf{77\%} & \textbf{700} \\
\bottomrule
\end{tabular}
\caption{Blind pairwise \emph{human} preference for \textsc{ToFu} by category, with 7 annotators $\times$ 100 cases in total.}
\label{tab:human-eval}
\end{table}

\section{UI demonstration}

\begin{figure*}[!htbp]
  \includegraphics[width=0.49\linewidth]{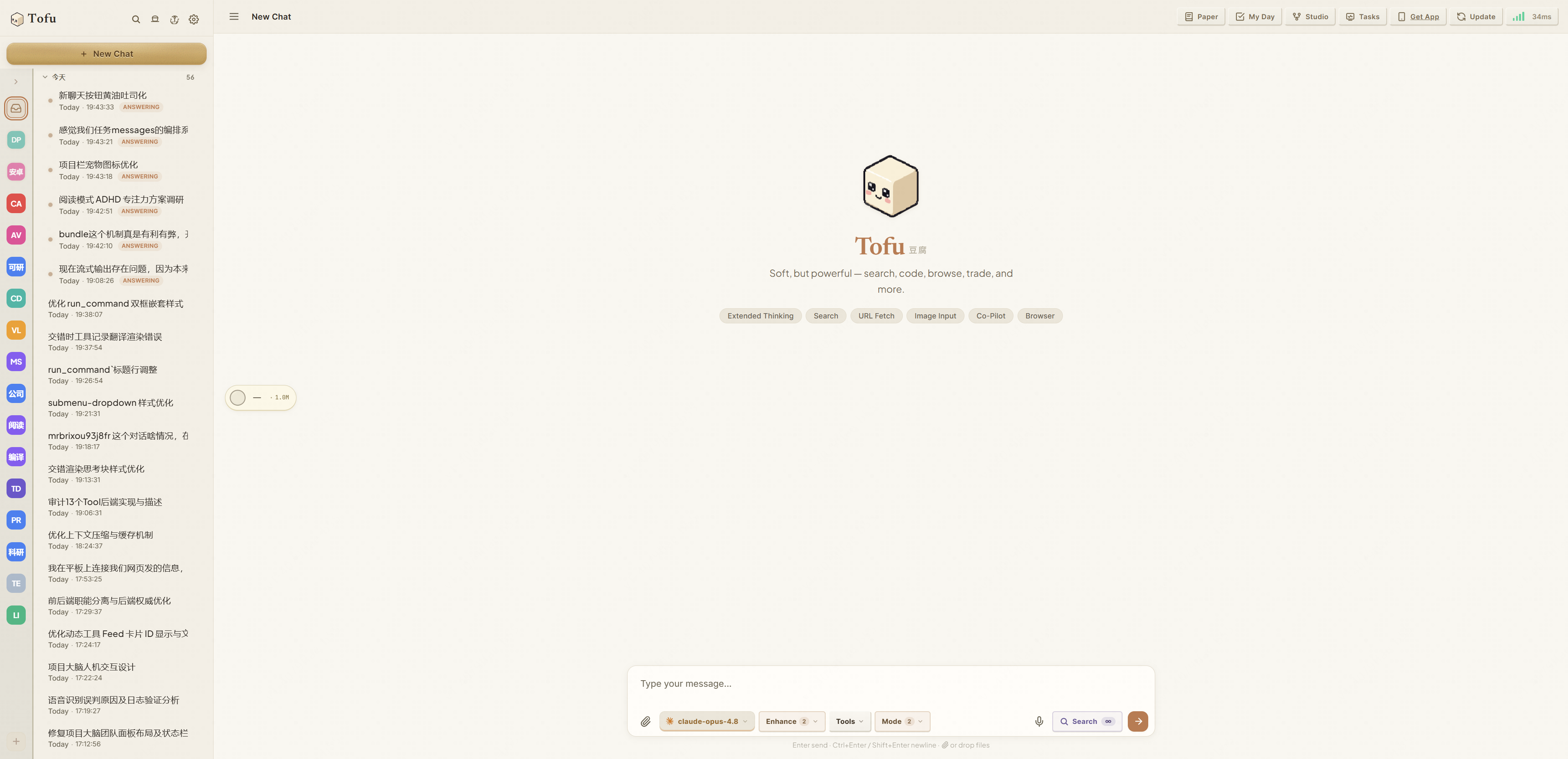} \hfill
  \includegraphics[width=0.49\linewidth]{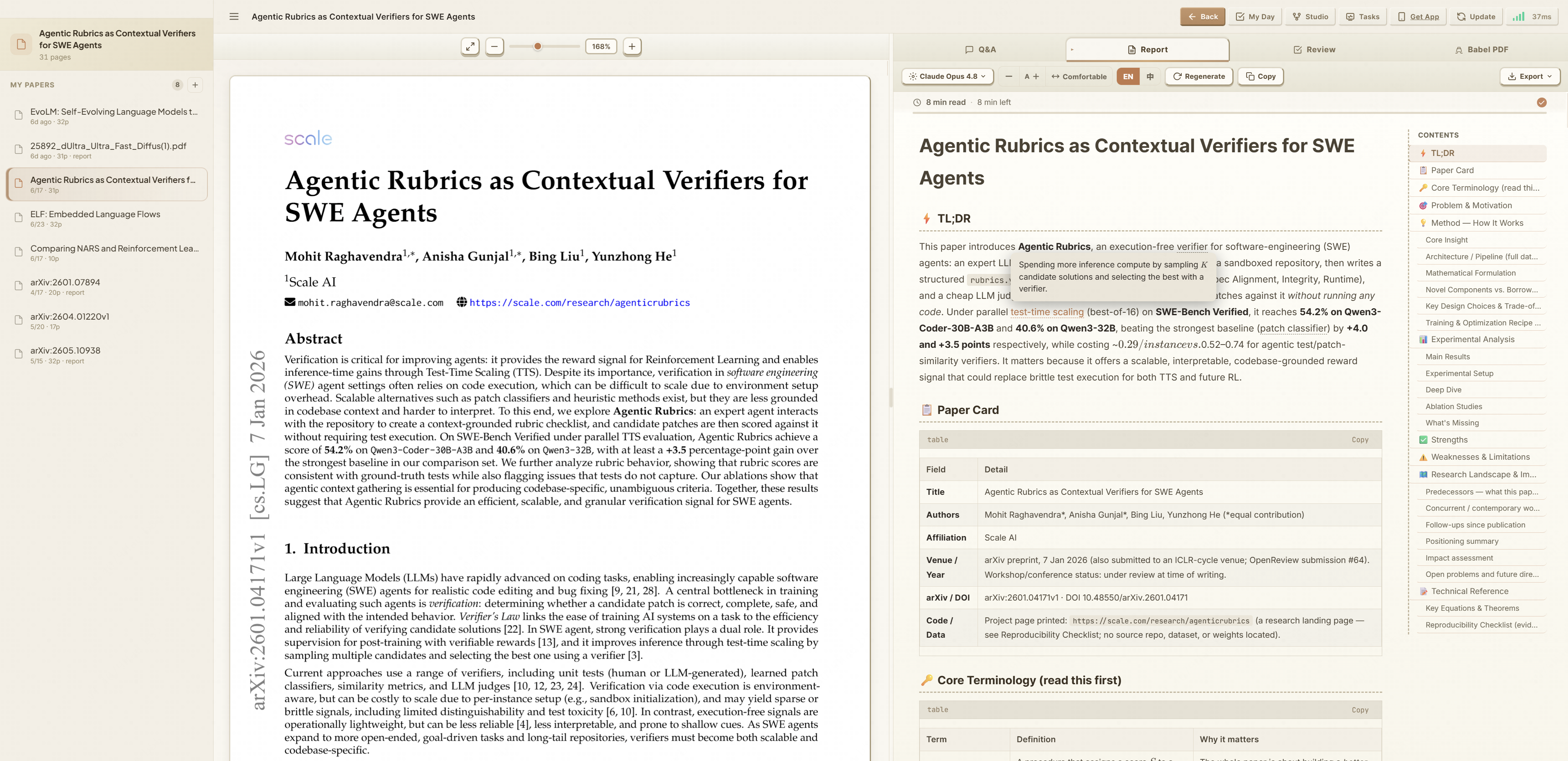}
  \caption {Demonstration UI page of \textsc{ToFu} and paper reader.}
  \label{fig:interface}
\end{figure*}

Fig. \ref{fig:interface} visualize the demonstration UI pages of \textsc{ToFu} and paper reader, respectively.

\section{Case study: Booking a flight}

We include one complete, unedited trace to show that \textsc{ToFu}'s
tool-call loop, the same one used on SWE-bench, transfers to everyday
web interaction, and that the run is fully attributable end to end.

\paragraph{Task.} \emph{``Find flights from Washington to New York
tomorrow between 13:00--18:00 on trip.com, shortest duration first.''}
\textsc{ToFu} resolves ``tomorrow''$\to$2026-06-29 and the cities to
airport groups (\texttt{WAS}, \texttt{NYC}) with no further input.

\paragraph{Trace.} Listing~\ref{lst:trip-trace} is the full five-round
trace: inspect the session, build a parameterized search URL, wait for
the page to settle, screenshot for grounding, then inject the round-5
DOM probe in Listing~\ref{lst:trip-js}, which filters candidate nodes by
a time-and-stops pattern and de-duplicates them into machine-readable rows.

\begin{lstlisting}[style=tofucode,caption={Trace of 5 rounds tool-call.},label={lst:trip-trace}]
[R1] browser_list_tabs
  args:   {}
  result: 3 tabs open; none on trip.com.

[R2] browser_create_tab
  args:   {"active": true, "url":
   "https://us.trip.com/flights/showfarefirst
    ?dcity=was&acity=nyc&ddate=2026-06-29
    &triptype=ow&class=y&quantity=1
    &searchboxarg=t"}
  result: Created tab #12166360.

[R3] browser_wait
  args:   {"tab_id": 12166360, "time": 6}
  result: Waited 6.0s.

[R4] browser_screenshot
  args:   {"tab_id": 12166360,
           "full_page": false}
  result: Screenshot captured (png).

[R5] browser_execute_js
  args:   {"tab_id": 12166360, "code": <see Listing 2>,
   "description": "Extract all flight rows
    (times, duration, stops, airports, price)"}
  result: 40 text rows -> 12 in-window flights.
\end{lstlisting}

\begin{lstlisting}[style=tofucode,language=,caption={Core of the round-5 DOM probe.},label={lst:trip-js}]
Array.from(document.querySelectorAll('[class*="flight-item"],div'))
  .filter(el => /\d{1,2}:\d{2}\s?(AM|PM)/.test(el.innerText)
             && /\dh|Nonstop|stop/i.test(el.innerText)
             && el.innerText.length < 400)
  .map(el => el.innerText.replace(/\n+/g,' | ').trim());
\end{lstlisting}

\paragraph{Result.} Of the scraped rows, twelve are nonstop and inside
the 13:00--18:00 window. Sorted by duration, the two shortest tie at
\textbf{1h22m}: United (Gojet) \texttt{DCA}$\to$\texttt{EWR} at \$215,
departing 1:00\,PM and 3:10\,PM. \textsc{ToFu} surfaces these and states
its assumptions explicitly (all-airport \texttt{WAS}/\texttt{NYC};
prices unverified, nothing booked). This visible chain of resolve,
navigate, ground, scrape, filter, sort, and recommend-with-caveats is
exactly the transparency the white-box design targets.

\section{Why \textsc{ToFu} outperforms OpenCode}
\label{sec:design-cmp}
Table~\ref{tab:design-cmp} contrasts the two designs across
context, tools, orchestration, and recovery. The decisive axis is recovery:
OpenCode does not retry by default and ends the turn on an interrupted stream,
so one transient failure strands the instance, whereas \textsc{ToFu} turns such
failures into in-loop retries rather than dead runs.

\begin{table*}[htbp]
    \centering
    \small
    \setlength{\tabcolsep}{6pt}
    \renewcommand{\arraystretch}{1.2}
    \begin{tabular}{@{}>{\raggedright\arraybackslash}p{2.35cm} >{\raggedright\arraybackslash}p{6.15cm} >{\raggedright\arraybackslash}p{6.15cm}@{}}
    \toprule
    \textbf{Design axis} & \textbf{OpenCode} & \textbf{\textsc{ToFu}} \\
    \midrule
    \multicolumn{3}{@{}l}{\textbf{Context management}}\\[2pt]
    Long tool outputs & Every result is hard truncated to a fixed size, and the overflow is saved to a file the agent must re-open & Keeps a bounded preview plus a re-expandable reference, and leaves file reads intact to avoid re-read loops \\
    Reusing old context & Rewrites old tool results in place, which invalidates the provider prompt cache for later turns & Replaces only cold results with short placeholders while keeping a hot tail, so the cached prefix is never rewritten \\
    Cache breakpoints & Placed by position on the first and last messages, recomputed every turn, and the pruning step mutates the prefix they depend on & Of the four markers, one is reserved for the growing tail and one for the last tool, so system blocks cannot crowd them out; the stable prefix is cached for one hour and the tail for five minutes \\
    When context is full & A single fixed threshold triggers one summarization pass & A zero cost pass runs every round, and an LLM summary fires only above about 80\% of the limit \\
    Summary focus & The same generic prompt regardless of the current task & Rates each past turn for relevance to the current query, keeping critical turns verbatim and dropping the rest \\
    Persistent memory & None, only human written instruction files and the raw conversation history & Markdown records the agent writes itself, retrieved by keyword and surfaced proactively across sessions \\
    Change reminders & Only mode switch notices, edited into an existing message & A recently-modified-files and next-step reminder, added as a new message so the prefix stays intact \\[4pt]
    \multicolumn{3}{@{}l}{\textbf{Tool design}}\\[2pt]
    Multi file operations & One file per call, so reading or editing $N$ files needs $N$ separate calls & Batched, so one call reads or edits many files, amortizing per call overhead and reducing turns \\[4pt]
    \multicolumn{3}{@{}l}{\textbf{Orchestration}}\\[2pt]
    Reasoning loop & A single agent calls tools in one flat loop, with no review step & A Planner, Worker and Critic loop with a verification gate \\
    Sub agent execution & One shot delegate and wait, run one at a time with no dependencies & A dependency aware graph runs independent agents in parallel and defers dependent ones \\
    Agent communication & Results return only to the parent, so sibling agents cannot share findings & Each finished result is injected into its dependents, and a coordinator can spawn follow up agents from intermediate findings \\[4pt]
    \multicolumn{3}{@{}l}{\textbf{Robustness (unattended runs)}}\\[2pt]
    Failed request & No automatic retry by default, and no switch to another provider or key & Bounded automatic retry, model fallback, and provider or key rotation \\
    Interrupted stream & Ends the turn & Re-runs the interrupted turn, up to 16 times with backoff \\
    Failed tool call & Marked as an error & Returned to the model as a structured request to retry with corrected input \\
    \bottomrule
    \end{tabular}
    \caption{Source-verified design comparison of \textsc{ToFu} and OpenCode.}
    \label{tab:design-cmp}
\end{table*}

\end{document}